\documentclass{statsoc}

\usepackage[a4paper]{geometry}
\usepackage{graphicx}
\usepackage[textwidth=8em,textsize=small]{todonotes}
\usepackage{amsmath}
\usepackage{natbib}
\usepackage{amsfonts}
\usepackage{float}
\usepackage{xcolor}
\usepackage{hyperref}

\newcommand{\norm}[1]{\left\lVert#1\right\rVert}
\newcommand{\bfs}{{\mathbf{s}}}
\newcommand{\bfx}{{\mathbf{x}}}
\newcommand{\bfa}{{\mathbf{a}}}
\newcommand{\bfy}{{\mathbf{y}}}
\newcommand{\bfw}{{\mathbf{w}}}
\newcommand{\bfd}{{\mathbf{d}}}
\newcommand{\bfv}{{\mathbf{v}}}
\newcommand{\bfX}{{\mathbf{X}}}

\newcommand{\calX}{{\mathcal{X}}}
\newcommand{\calY}{{\mathcal{Y}}}
\newcommand{\calA}{{\mathcal{A}}}
\newcommand{\loss}{\text{loss}}
\newcommand{\R}{{\mathbb{R}}}

\DeclareMathOperator*{\argmax}{arg\,max}

\title[]{Estimating a Brain Network Predictive of Stress and \\ Genotype with Supervised Autoencoders}
\author[Author 1 {\it et al.}]{Austin Talbot}
\address{Pillar Biosciences Inc.,
Natick MA,
USA.}
\email{talbota@pillarbiosci.com}
\author{David Dunson}
\address{Duke University,
Durham,
USA.}
\author{Kafui Dzirasa}
\address{Duke University,
Durham,
USA.}
\author{David Carlson\footnote{Correspondence should be addressed to: david.carlson@duke.edu; Hudson Hall, 100 Science Dr., Durham, NC 27710}}
\address{Duke University,
Durham,
USA.}
\author{}
\begin{document}
\begin{abstract}
Targeted stimulation of the brain has the potential to treat mental illnesses. We propose an approach to help design the stimulation protocol by identifying electrical dynamics across many brain regions that relate to illness states. We model multi-region electrical activity as a superposition of activity from latent networks, where the weights on the latent networks relate to an outcome of interest. In order to improve on drawbacks of latent factor modeling in this context, we focus on supervised autoencoders (SAEs), which can improve predictive performance while maintaining a generative model.  We explain why SAEs yield improved predictions, describe the distributional assumptions under which SAEs are an appropriate modeling choice, and provide modeling constraints to ensure biological relevance of the learned network. We use the analysis strategy to find a network associated with stress that characterizes a genotype associated with bipolar disorder. This discovered network aligns with a previously used stimulation technique, providing experimental validation of our approach.

\textit{Keywords: } Dimensionality Reduction; Factor Analysis; Joint Model; Neuroscience; Supervised Autoencoders

\end{abstract}

\section{Introduction}

Mental illnesses are among the most debilitating medical conditions due to their prevalence \citep{Lewinsohn1998} and the vast toll they take on individuals \citep{Monkul2007} and society \citep{Leon1998}. Yet the brain and the malfunctioning neural dynamics responsible for mental illness remain largely a mystery. Because of this, the National Institutes of Health is actively encouraging research \citep{Insel2013} to understand the brain at all levels of organization. Of the many types of electrophysiology commonly studied, our work focuses on understanding the dynamics of region-specific voltages in the brain, known as Local Field Potentials (LFPs). Field potentials have proven to be good candidates for understanding dynamics because the learned features are predictive of neuropsychiatric conditions \citep{Veerakumar2019} and can generalize to new individuals \citep{Li2017}, allowing us to characterize and predict diseases using features that apply to the general population rather than a specific individual.

Recent developments have allowed neuroscientists to test hypotheses on neural dynamics by directly modifying the brain dynamics underlying actions and behaviors. For example, optogenetics \citep{Boyden2005} is a set of techniques that use a virus to modify the cells in specific brain regions to respond to light by firing action potentials. The neural dynamics can then be modified by shining light through implanted fiber optic cables into these modified regions, inducing aggregate cell behavior that in turn affects the LFPs \citep{Buzsaki2012}. The benefits of this development cannot be overstated. Neuroscientists can now use experimental manipulation to argue for the causality of cell types in behavior, rather than mere correlation. But more importantly, it opens the door for targeted neurostimulation treatments of mental illnesses. Many of the current treatment methods for psychiatric disorders rely on drugs, which can have debilitating side effects \citep{Short2018} due to their broad distribution in the brain and the low temporal resolution at which they impact brain cells. Direct manipulation of the faster dynamics responsible for the illness has the potential to cure these diseases without the negative consequences of medication.

However, we must move beyond mere predictive models to obtain the necessary insight into causal mechanisms for such interventions to succeed. For stimulation to be effective we must find a neural circuit responsible for the observed dynamics \citep{Kravitz2010} rather than just finding predictors that correlate with the behavior. Such collections of neural circuits, heretofore referred to as networks, inferred by a statistical method must be biologically relevant and provide enough information about brain function to find targets for stimulation. This is analogous to creating a harmony in an orchestra: one must be able to identify the out-of-tune instrument rather than merely detecting the errant notes produced.

This analogy of an orchestra is particularly apt for neuroscience where it is commonly assumed that a small number of these unobserved networks give rise to the high-dimensional dynamics \citep{Medaglia2015}. This viewpoint has often led researchers to use factor analysis as a method for understanding brain activity networks. Many factor models have been created to incorporate varying assumptions about the latent networks and resulting dynamics, including independent component analysis (ICA) \citep{Du2016}, non-negative matrix factorizations (NMF) \citep{Lee1999}, and principal component analysis (PCA) \citep{Feng2015}. Some factor models have been developed to specifically identify possible locations where stimulation would be effective \citep{Gallagher2017}.

Our objective is to find a brain network in a mouse model associated with two traits: being in a stressful state and having a genotype that has been linked to bipolar disorder, Clock-$\Delta$19, compared to wild-type littermate controls using the data described by \citet{Carlson2017}.  Mice with the Clock-$\Delta$19 genotype have been shown to be resilient to stressful situations \citep{Dzirasa2011a,Sidor2015,Roybal2007}, so understanding how mice with and without the genotype differ can assist in understanding of stress response networks.  The data in this setting are LFP recordings from both Clock-$\Delta$19 and wild-type mice in a variety of situations ranging from not stressful to highly stressful. It is often a scientific goal to find a single network that differentiates between mice in different groups. This goal necessitates that the predictive information is located in a single factor, and this factor must be interpretable to neuroscientists in order to find candidate locations for stimulation. Manipulation of the dynamics will not be possible unless all these conditions are met. 

Factor models that focus exclusively on modelling the electrophysiology data often fail to find networks that are predictive of traits of interest (i.e., stress condition, genotype). This suggests using joint factor models that assume common latent variables underlying both electrophysiology and traits. Unfortunately, joint factor models place a very strong prior on the predictive coefficients that can substantially degrade predictive performance \citep{Hahn2013}, particularly if the number of estimated factors is smaller than the number of true factors. The true number of networks responsible for neural dynamics are numerous, certainly more than the number of factors we can reliably estimate. In addition, most of these networks are unrelated to the traits of interest. These irrelevant networks can be particularly dominant, for example those related to motion \citep{Khorasani2019} or blinking \citep{Joyce2004}.  One potential solution is to increase the weight on the predictive component when fitting the joint model.  However, we find that although such an approach can do well in-sample, it fails to accurately infer predictive networks for test subjects based on information on electrophysiology data alone and hence cannot address our goals.

Supervised autoencoders (SAEs) \citep{Ranzato2008} have arisen as an alternative to classical joint factor models. An autoencoder is a method for dimensionality reduction that maps inputs to a lower dimensional space using a neural network known as an encoder. The latent space is then used to reconstruct the original observations by a neural network called a decoder \citep{Goodfellow2017}. Both the encoder and the decoder are learned simultaneously to minimize a reconstructive loss. Supervised autoencoders add another transformation from the latent factors to predict the outcome of interest, thus encouraging the latent factors to represent the outcome well \citep{Xu2017}. These have been used with great success in image recognition, especially in the context of semi-supervised learning \citep{Le2018,Pezeshki2016,Pu2016}.
In this work, we show how SAEs can be adapted to solve a difficult problem in neuroscience: finding a single network suitable for stimulation to modify a trait. Finding this predictive network has been a difficult problem in neuroscience, as experimental predictive ability has failed to match theoretical expectations with traditional inference techniques. We show empirically on synthetic data that model misspecification in generative models is a substantial contributor to these difficulties. We then demonstrate on both synthetic data and our experimental dataset that our SAE-based approach is able to successfully identify such a predictive network, even under substantial misspecification. Interpreting this network leads to natural conclusions on potential stimulation targets. Previous studies have experimentally modified this target to successfully modulate animal behavior \citep{Carlson2017}. Together, these  contributions provide substantial evidence for the promise of our method as a useful tool for experimentalists to develop stimulation methods as potential treatments for mental illness.  \textcolor{black}{Notably, the proposed methodology has been used to successfully design a targeted neurostimulation protocol \citep{mague2022brain}, providing further evidence of this claim}.


This paper is organized as follows: Section 2 contains a description of the data and motivation, {\color{black}along with defining joint factor analysis and demonstrating drawbacks under model misspecification}.  Section 3 derives our SAE approach while considering some basic properties and issues to motivate modifications of previous SAE frameworks. Section 4 provides two synthetic examples to illustrate the benefits of SAEs, one with a standard NMF model and another using synthetic LFPs. Section 5 shows that our approach learns predictive networks of genotype and stress, describes how the inferred ``stress network'' could be modified through stimulation, and relates the network to previous literature.  In Section 6 we provide concluding remarks and disucss further directions for research. 

\section{Data and model}

{\color{black} In this section, we introduce the electrophysiological data analyzed in this paper and the scientific motivation for a latent variable model.  We highlight the need for a single latent variable to predict the experimental outcomes; here, these outcomes relate to an animal model of stress and a genotype associated with bipolar disorder. As part of this section, we highlight the importance of robustness to certain types of model misspecification (here, primarily latent dimensionality), which are nearly ubiquitous in neuroscience. Unfortunately, typical inference strategies for latent variable models are not robust to such misspecification, as we show in  simulations. We show misspecification has a strong deleterious effect on previous approaches to learn predictive factor models.}

\subsection{Electrophysiology: The Tail Suspension Test}

\begin{figure}
\begin{center}
\includegraphics[width=\textwidth]{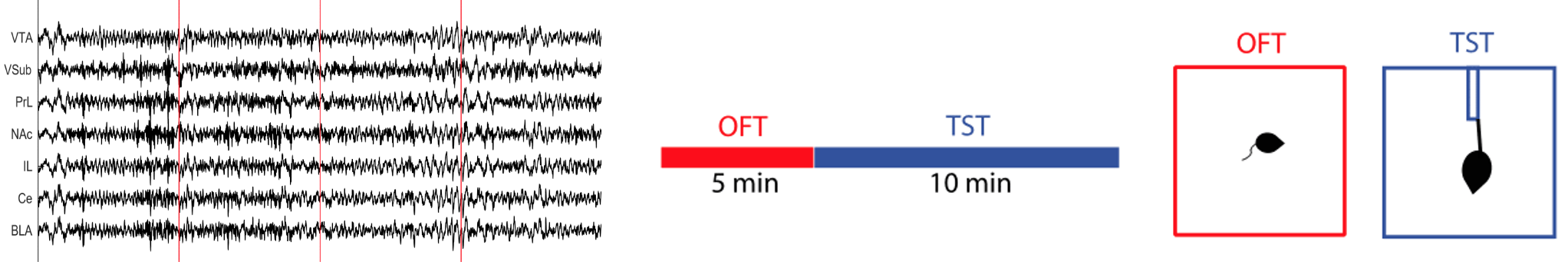}
\end{center}
\caption{\label{fig:experiment} On the left we show example LFP recordings in a subset of the regions measured. The LFPs were divided into one-second windows to yield discrete observations. On the right we provide a simple diagram of the TST experiment. The mice were recorded in the home cage as a non-stressful environment. They were then recorded for 5 minutes in a mildly stressful open field test (OFT) and then 10 minutes in a highly stressful tail suspension test (TST).}
\end{figure}

The LFPs analyzed in the Tail Suspension Test (TST) came from 26 mice of two genetic backgrounds (14 wild type and 12 Clock-$\Delta$19). The Clock-$\Delta$19 genotype has been used to model bipolar disorder \citep{VanEnkhuizen2013a}. Each mouse was recorded for 20 minutes across 3 behavioral contexts: 5 minutes in its home cage (non-stressful), 5 minutes in an open field (arousing, mildly stressful), and 10 minutes suspended by its tail (highly stressful). Data were recorded from 11 biologically relevant brain regions with 32 electrodes (multiple electrodes per region) at 1000 Hz. These redundant electrodes were implanted to allow for the removal of faulty electrodes and electrodes that missed the target brain region. We chose to average the signals per region to yield an 11-dimensional time series per mouse. This was done because the brain region represents the smallest resolvable location when modeling multiple animals; multiple electrodes function as repeated measurements and averaging allows us to reduce the variance of the measured signal. A visualization of these data and the experimental design can be seen in Figure \ref{fig:experiment}.

We want to determine a single brain network that predicts stressful activity, so we consider all data from the home cage as the negative label (stress-free condition) and all data from the other two conditions as the positive label (stressed condition). A second prediction task is to determine brain differences between the genetic conditions (i.e., what underlying differences are there between the wild type and bipolar mouse model?). There is strong evidence to support the belief that the behaviorally relevant aspects of electrophysiology are frequency-based power within brain regions, as well as frequency-based coherence between brain regions \citep{Hultman2016,Uhlhaas2008}. We discretized the time series into one-second windows to model how these spectral characteristics change over time. Windows containing saturated signals were removed (typically due to motion artifacts). While there are methods that can characterize the spectral features on a continuous time scale \citep{Prado2010}, the behavior we deal with changes over a longer scale than the observed dynamics. Consequently, it is more effective to discretize and obtain sharper spectral feature estimates \citep{Cohen1995} that are more amenable to factor modelling.

\begin{figure}
\begin{center}
\includegraphics[width=\textwidth]{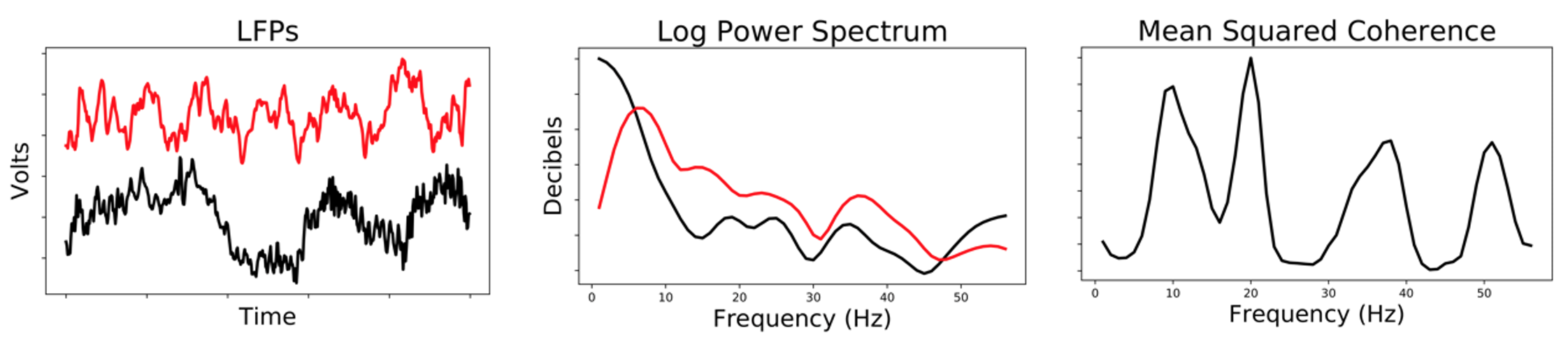}
\end{center}
\caption{\label{fig:features} A visualization of the features used for CSFA-NMF. On the left we show a 1-second window for the LFPs in two brain regions. The middle shows the power spectra for each LFP from 1-56 Hz estimated using Welch’s method. The right shows the mean squared coherence, which we used to quantify the coherence between the two LFPs. We show two channels for clarity; in practice we have many channels and calculate all pairwise coherences.}
\end{figure}

We chose to extract the relevant quantities from the recorded data prior to modeling rather than extracting spectral features in the modelling framework for simplicity; the extra modeling step would substantially increase the number of parameters in the model. The features related to power were computed from 1 Hz to 56 Hz in 1 Hz bands using Welch’s method \citep{Welch1967}, which is recommended in the neuroscience literature \citep{Kass2014}. We chose 56 Hz as a threshold to avoid the substantial noise induced at 60 Hz from the recording equipment, as prior literature has demonstrated that much of the meaningful information is contained in the lower frequencies. We calculated mean squared coherence at the same frequency bands for every pair of brain regions \citep{Rabiner1975}. This procedure converted each window of data into 3696 non-negative power and coherence features. Figure \ref{fig:features} shows two LFPs and the associated features calculated from the recordings.

\subsection{Difficulties with Joint Factor Models}

{\color{black}

After extracting the relevant quantities described above, we have high-dimensional data (which scales by the square of the brain regions and number of frequencies considered). A natural approach for characterizing the data is to use latent variable models. These models posit that an unobserved latent space (factors or in the neuroscience field ``networks'') generates the observed dynamics, with the mapping between the two (loadings) explaining the correlations in the observed covariates. As the dimensionality of the latent space is often far lower than the observed space, this allows for efficient representation of the correlations in the data \citep{Bishop2006}. The covariates in neuroscience often have strong correlations, so it is unsurprising that latent variable models have been extensively used \citep{bassett2017network}. The latent representation must also contain information relevant to the outcomes relating to stress and genotype.} 

There is a vast literature of methods that reduce dimensionality while retaining predictive information. One widely studied approach is sufficient dimensionality reduction (SDR), which exploits the assumption that the response is conditionally independent of the predictors given a projection of the predictors into a lower dimensional subspace. Examples of this technique include sliced inverse regression \citep{Li1991} and extensions \citep{Fukumizu2009,Wu2009}. However, these approaches require stringent assumptions on the data and the outcome, which limits their utility in neuroscience. In particular, these models induce a strong prior on the predictive coefficients, which leads results to be sensitive to model misspecification. This difficulty can be seen particularly clearly with misspecification of the latent dimensionality \citep{Hahn2013}.

{\color{black} Joint factor models are appealing modelling choices to incorporate information about both the covariates and the outcomes, as demonstrated by the popularity and use of supervised probabilistic PCA \citep{Yu2006}, adding supervised variables in highly-related topic models \citep{mcauliffe2007supervised}, and the use of the concept in deriving supervised dictionary learning algorithms \citep{Mairal2009}. We now formally introduce these models. To define our notation, let  $\{\bfx_i\}_{i=1,\dots,N}\in\calX$ be the measured dynamics (predictors) and $\{y_i\}_{i=1,\dots,N}\in\calY$ be the corresponding trait or outcome. The latent factors are denoted $\{\bfs_i\}_{i=1,\dots,N}\in\mathbb{R}^L$, $\Theta$ are the parameters relating the factors to $\bfx$, and $\Phi$ are the parameters relating the factors to $y$. To handle intractable integrals involved in marginalizing over the latent factor distributions  \citep{Gallagher2017}, two common approaches are to optimize a variational lower bound similar to a variational autoencoder \citep{Kingma2013} or alternatively to estimate the factors jointly with the model parameters \citep{Mairal2009}. We choose to proceed with the latter, due to its previous usage in similar applications.

Given this strategy, a generic objective function for a joint factor model is
\begin{equation}\label{eq:jfm}
    \max_{\Phi,\Theta,\{\bfs_i\}_{i=1,\dots,N}}\sum_{i=1}^N\mu \log p(\bfx_i|\bfs_i,\Theta) + \log p(y_i|\bfs_i,\Phi) + \log p(\bfs_i) + \log p(\Theta) + \log p(\Phi)
\end{equation}
The tuning parameter $\mu$ controls the relative weight between  reconstruction and supervision. Including $\mu$ is somewhat atypical, and we note that setting $\mu=1$ recovers the log-likelihood of a standard joint factor model. In practice, it may be important to modify $\mu$ to upweight the importance of prediction by setting $\mu\ll1$. This term can be thought to correspond to a fractional posterior $p(\bfx|\Theta)^\mu p(y|\bfx)$, which has been previously used in robust statistics \citep{bhattacharya2019bayesian}, and is highly related to the common practice of modifying the noise variances on $\bfx$ and $y$ in joint factor models \citep{Yu2006}.  This tuning parameter is frequently used in machine learning \citep{Mairal2009}. Given this objective function, we find the optimal values of the latent factors, $\Theta$, and predictive coefficients $\Phi$. This is commonly done with stochastic methods due to their computational advantages in large datasets.} 

\begin{figure}
\begin{center}
\includegraphics[width=\textwidth]{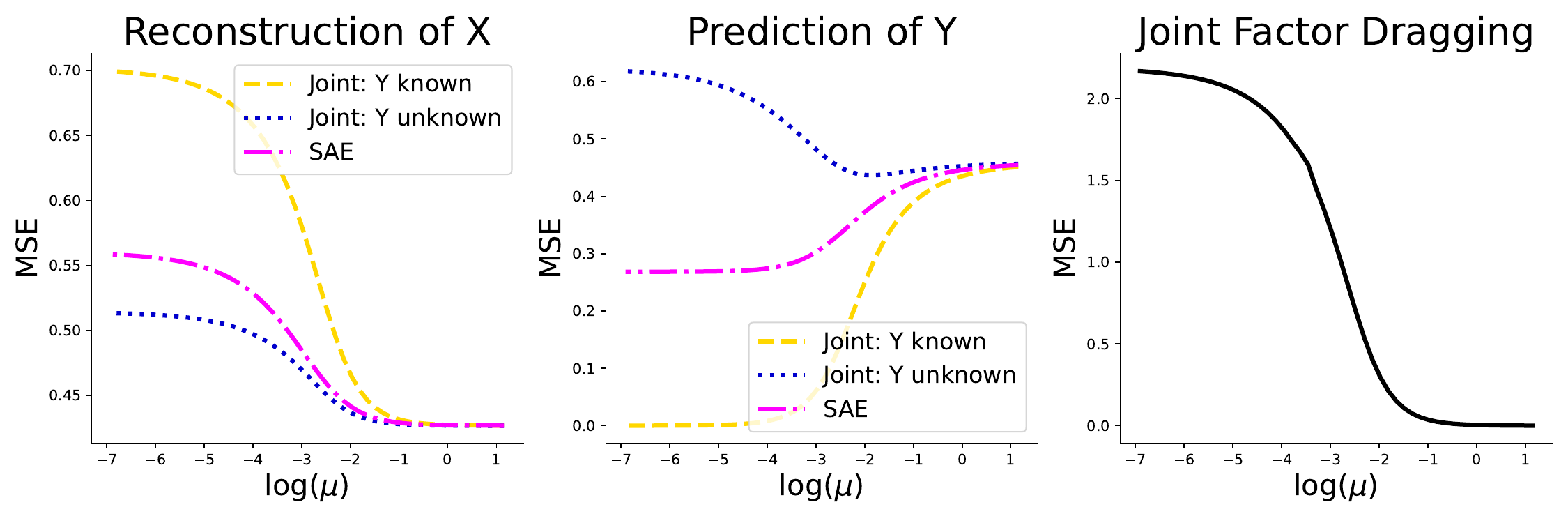}
\end{center}
\caption{\label{fig:dragging} There are 3 true latent factors but the model has 2 latent factors. We vary the supervision strengths ($\mu$) over a range of values and report the MSE for three cases: a joint model where the factors are estimated with the trait known (Joint: Y known), a joint model where the trait is unknown (Joint: Y unknown), and a supervised autoencoder (SAE).}
\end{figure}

However, joint models are not without their own drawbacks, as we now demonstrate using analytic solutions with an $L_2$ loss on a misspecified model. Furthermore, as motivation for our approach, we show that SAEs are not as affected by this type of misspecification. We generated a synthetic dataset corresponding to supervised probabilistic PCA \citep{Yu2006}, a simple joint linear model. We set the number of predictors to 20 with a single outcome and a latent dimensionality of 3. The largest factor was unpredictive of the outcome. Additionally, we limited the inferred dimensionality to 2. This mimics our statistical models that may under-specify the true number of brain networks. We fit a linear joint model and an SAE, {\color{black} which we will fully describe in Section 3}, with 2 components over a range of supervision strengths using the analytic solutions and show the results in Figure \ref{fig:dragging}. At low supervision strengths (large values of $\mu$) both the joint model and SAE focus on reconstructing the data. At higher strengths (small values of $\mu$) they sacrifice some of the reconstruction for better predictions of the outcome. However, the joint model simply overfits the factors by making them overly dependent on $y$, leading to {\color{black} an effect we refer to as ``factor dragging''. We define this as 
\begin{equation}
    FD = \frac{1}{N}\sum_{i=1}^N |g(\bfx_i,y_i)- f(\bfx_i)|
\end{equation}
where $g(\cdot)$ and $f(\cdot)$ represent the mapping to $s$ when both $\bfx$ and $y$ are observed and when just $\bfx$ is observed, respectively. This represents the difference between the factor estimates when the outcome is known (training) and when the outcome is unobserved (testing).  For a joint factor model in this setup, this corresponds to $g(\bfx_i,y_i)=\argmax_\bfs \mu \log p(\bfx_i|\bfs,\Theta)+\log p(y_i|\bfs,\Phi)+\log p(\bfs)$ and $f(\bfx_i)=\argmax_\bfs \mu \log p(\bfx_i|\bfs, \Theta) +\log p(\bfs)$.
 When the latent space is dedicated to reconstructing $\bfx$, this difference will be small.  However, as $\mu$ is increasingly small, the values of $\bfs$ will be largely influenced by the predictive loss on $y$ when it is observed, and this difference will become increasingly substantial when $y$ is not observed. This is particularly relevant to our applications, as the dimensionality of $\bfx$ often corresponds to between 3000-10000 features. In order to influence the latent space we often must roughly ``balance'' the reconstructive loss and predictive loss, corresponding to a value of $\mu\approx1/3000$. The rightmost plot of Figure 3 shows this pattern, which is increasingly large with strong supervision. Large discrepancies will lead to poor predictions, as it indicates that the absence of knowledge of the outcome $y$ (inherent to prediction) dramatically affects the latent representation, and by extension the reconstruction/prediction. While some efforts have been made to overcome these known issues, particularly Task-driven Dictionary learning \citep{mairal2011task}, this does not solve the problem and lacks the broad applicability to the many generative models used in neuroscience. In contrast, our SAE approach, which uses the same map from $\bfx$ regardless of whether $y$ is observed, circumvents these issues and is broadly applicable as we fully explain in Section 3.} 

There have been efforts to fix this problem, such as ``cutting the feedback'' approaches \citep{Liu2009,McCandless2010,Plummer2014}. In MCMC-based methods with this approach, the samples of the latent variables are drawn using only the electrophysiology (hence the influence from the outcome is cut). The corresponding maximum likelihood technique fits a factor model exclusively on the electrophysiology and then fits a predictive model using the estimated factors. The problem of cutting the feedback is immediately apparent; dominant networks are often associated with motion, maintaining homeostasis, or other irrelevant behaviors. The latent space will likely not contain information relevant to prediction when the outcome is ignored \citep{Jolliffe1982}. 

An alternative solution for factor models is to substantially increase the number of latent components. This will capture the relevant information, but the information will be scattered across multiple networks. {\color{black} This is particularly harmful in our application, as the latent network itself is used to choose stimulation targets \citep{Carlson2017}. If the predictive information is located in a single network, neuroscientists can choose regions that are active and well-connected in the predictive network to stimulate or run downstream tests on a single variable with higher statistical power. If the generative model is a good representation of the neural dynamics, this will result in a global stimulation of the predictive network resulting in altered behavior. However, if the predictive information is scattered across multiple networks, it is impossible to identify which connected regions of the networks will result in altered behavior.} We will show that all these issues arise when predicting stress and genotype using electrophysiology on the TST dataset, and that our SAE-based approach {\color{black} addresses these issues, resulting in} substantially improved performance.

\section{An SAE-based Approach}

{\color{black} We now introduce a novel SAE-based approach to learn a single predictive latent factor, and discuss reasons for superior performance in practice. We then derive analytic solutions for SAEs in a linear model. These solutions are used to contrast the behavior of SAEs to traditional inference techniques in an illustrative example. Finally, we analyze appropriateness of SAEs in biological applications based on common assumptions.}

\subsection{Why do SAEs Yield Superior Predictions Compared to Joint Factor Models?}

Given a new sample $\bfx^*$, prediction of $y^*$ given $\bfx^*$ is straightforward once $\Theta$ and $\Phi$ have been estimated; simply compute point estimates of the factors and outcome as $\hat{\bfs}^*=\arg\max\log p(\bfs|\bfx^*)$ and $\hat{y}^*=\arg\max\log p(y|\hat{s}^*)$. However, these predictions are often quite poor. This issue does not stem from the inference method, as the example from Section (2) maximized $\Theta$ and $\Phi$ directly and yet still yielded poor predictions. Instead, the issue is due to the model attempting to explain all the variance of the outcome using the latent factors \citep{Hahn2013}. If there is a high amount of weight on predictive performance, a joint model will discover that the best predictor of $y$ is $y$ itself. This influence is most keenly felt when $L$ is small, as it is difficult to force the model to prioritize prediction of y without increasing the dependence on y for proper estimation of $\bfs$.

While there are a plethora of methods for selecting $L$, such as cross-validation or carefully designed priors \citep{Bhattacharya2011}, these methods will be largely unhelpful in improving predictive performance. These methods value a parsimonious representation of the joint likelihood, corresponding to fewer latent factors. This conflicts with a prior focused on predictive ability which would encourage \textit{more} factors. While cross-validation of the predictive accuracy over the dimensionality seems like an appealing option, in practice it still yields subpar predictions as compared to a purely predictive model and has difficulty finding a single predictive latent factor.

SAEs are an alternative to joint models that modify a deep feedforward network to include an autoencoder in order to reconstruct the predictors. Let us define the predictive feedforward network as $\hat{y}=f_\Phi(\calA(\bfx))$, where $\calA:\calX\rightarrow\mathbb{R}^L$ is the encoder, and the reconstructive autoencoder is a composition of the encoder and decoder $\hat{\bfx}=g_\Theta(\calA(\bfx))$. For estimation, it is common to place losses on the reconstructive and predictive parameters as well as the latent factors as regularization. This yields an objective function,

\begin{equation}\label{eq:sae_loss}
    \min_{\Theta,\Phi,\calA}\sum_{i=1}^N \mu\loss(\bfx_i,\hat{\bfx}_i) + \loss(y_i,\hat{y}_i) +  \loss(\calA(\bfx_i)) + \loss(\Theta) + \loss(\Phi).
\end{equation}

To maintain the desired interpretability of joint models, we choose the parameters and losses to correspond to the parameters and negative log likelihoods from the joint model. {\color{black} An SAE learns a mapping from $\bfx$ to $\bfs$ that is used to predict y due to the supervision loss, incorporating the relevant information during training. However, this mapping is only dependent on $\bfx$ in contrast to joint models which depend on both variables as inputs. Once the SAE is estimated, at test time $\bfx$ alone is used in prediction and reconstruction of the electrophysiology.}  Because of this, SAEs are limited to modeling the variance in y that can be predicted by $\bfx$. From this point of view, the reconstruction loss can be considered a deep-learning version of regularization discussed in \cite{Hahn2013}, and was explicitly motivated as such in \cite{Le2018}. In these situations, the reconstruction loss is placed on the latent space functions as a method to reduce the complexity of the model, corresponding to a structural risk minimization (SRM) approach \citep{Vapnik1992}. This technique has been used successfully to regularize complex models \citep{Guyon1992,Kohler2002}. However, there is one substantial difference between our approach and a typical SRM approach. With an SRM, the complexity penalty shrinks with an increasing number of observations (corresponding to $\mu\rightarrow0$). In SAEs, $\mu$ is a fixed constant, as it is critical that the factors maintain biological relevance even with large numbers of observations. 

\subsection{When are SAEs an Appropriate Predictive Model?}

The previous section makes it clear that, relative to joint models, SAEs improve predictive performance and reduce the consequences of misspecification of the true latent dimensionality. However, it is difficult to intuitively see what these differences are, as most formulations of Equations (\ref{eq:jfm}) and (\ref{eq:sae_loss}) do not have analytic solutions and are learned with stochastic methods \citep{Bottou2018}. We develop novel analytic solutions when the likelihood is replaced with an $L_2$ loss. This yields a form similar to PCA, which can be considered the limiting case of probabilistic PCA as the variance of the conditional distribution vanishes \citep{Bishop2006}.

We assume that both the predictors and outcome are demeaned. For convenience, we will define matrix forms of the data and factors: $X=[\bfx_1,\dots,\bfx_N]\in\R^{p\times N}$,
$Y=[\bfy_1,\dots,\bfy_N]\in\R^{q\times N}$, and 
$S=[\bfs_1,\dots,\bfs_N]\in\R^{L\times N}$. 
We let the model parameters $\Theta=W$ and $\Phi=D$ for matrices
$W\in\R^{p\times L}$, 
$D\in\R^{q\times L}$. For the associated SAE model, 
we assume a linear encoder, $\calA(\bfx)=A\bfx$.

The solution for the concatenation of $W$ and $D$ of the joint factor model, $[W^T,D^T]^T$, can be found as an eigendecomposition of the matrix 

\begin{equation}\label{eq:mat_jm}
    B = \begin{bmatrix}
\mu XX^T & XY^T \\
\mu YX^T & YY^T
\end{bmatrix}
\end{equation}

When $Y$ is known, the factor estimates can be computed using the fixed-point equation detailed in the Appendix. When the outcome is unknown, the factor estimates are simply the projection of the data onto the latent space, $S=(W^T W)^{-1} W^T X$. This solution form is well-known from PCA of the predictors and outcome jointly, except for the minor extension with $\mu$. 

The solution for the concatenation of $W$ and $D$ under the SAE formulation as defined above is also an eigendecomposition with a slightly different matrix 
\begin{equation}\label{eq:mat_sae}
    B = \begin{bmatrix}
\mu XX^T & XY^T \\
\mu YX^T & YP_XY^T
\end{bmatrix}
\end{equation}
where $P_X=X^T(X^TX)^{-1}X$ is a projection matrix onto $X$. The estimate for $A$ can be computed via a fixed-point equation. We provide the details of the fixed-point equations and the derivations in the Appendix. The important aspect to note is that the SAE only models the variance of $\bfy$ that it is linearly predictable by $\bfx$ due to the term $P_X$. Thus, given that the latent space is only computed with $\bfx$, this formulation forces the model to find predictive factors using only the predictors.

We can use these analytic formulae to examine the distributional assumptions under which SAEs provide good predictive models. Previous work by \cite{Le2018} showed that a reconstruction loss added to a predictive model provided a bound on the generalization error in linear SAEs via sensitivity analysis \citep{Bousquet2002}. This corresponded to predictive gains with more common deep networks. However, bounds provided by sensitivity analysis can be loose \citep{Zhang2017} and ignore properties of the data generating process \citep{Bu2020}.
This led to a claim that reconstruction penalties empirically never harm performance.  While this is often true in practice, an intuitive counterexample to this claim is where the latent structure is not highly related to the outcome. The reconstruction loss will force the model to focus on high-variance predictors unrelated to the outcome, harming predictive performance. On the other hand, having the predictive information correlate with high variance latent factors yields substantial improvements, as estimates of high variance components can converge with small numbers of samples \citep{Shen2016}.  Therefore, it is important to evaluate whether a dataset is likely to benefit from an SAE-based approach before analysis.

\begin{figure}
\begin{center}
\includegraphics[width=\textwidth]{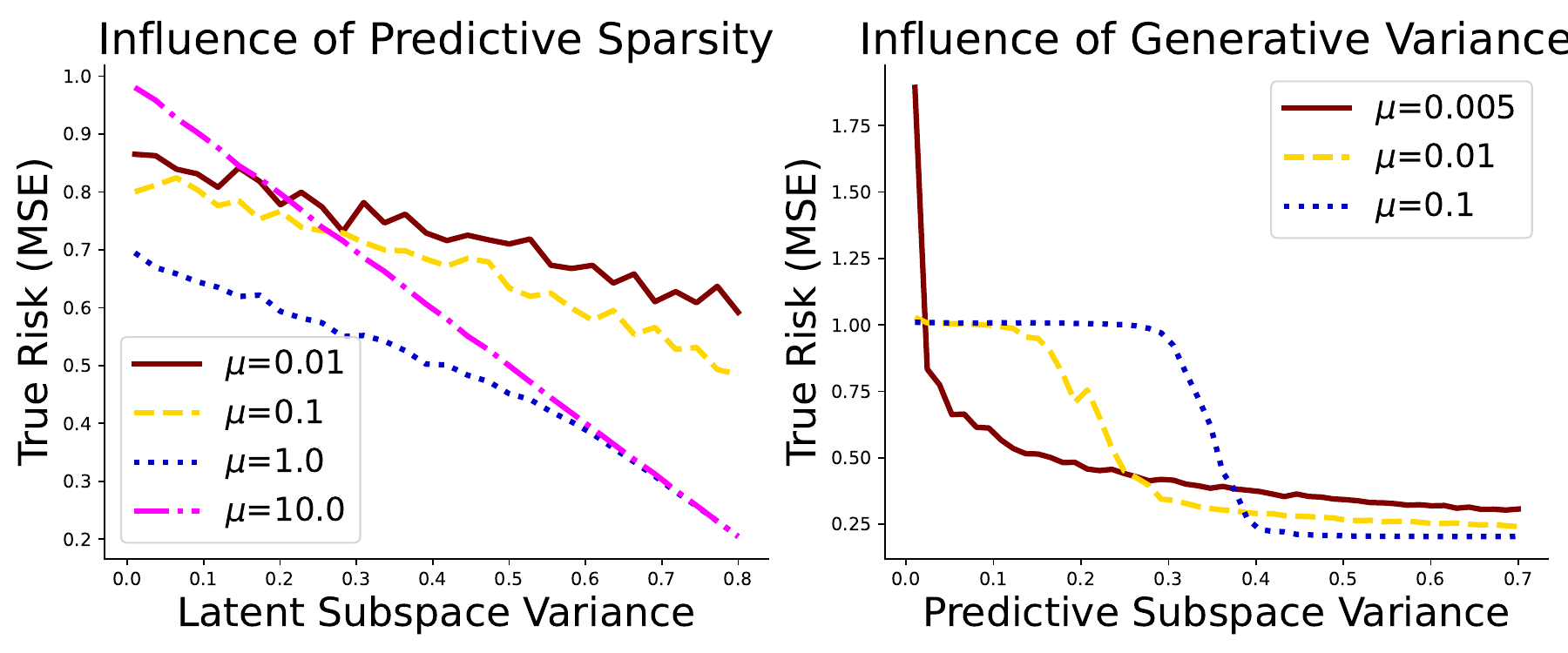}
\end{center}
\caption{\label{fig:distribution} The plot on the left shows effect of sparsity in dimensionality on the true risk (generalization error). The plot on the right shows how having predictive information align with high variance components of the covariates affects generalization error.}
\end{figure}

We empirically explore the distributional assumptions of SAEs on random matrices using the analytic solutions previously derived. Specifically, we show that SAEs are beneficial when the predictive information lies on a low dimensional manifold, especially one that correlates strongly with high-variance components. To demonstrate the first claim, we generated $200$ samples of predictors $\bfx\in\R^{300}$ with zero mean and unit variance with a single latent component $s\in\R$ explaining 70\% of the variance. The outcome $y\in\R$ was generated as $y_i=(1-\lambda)\bfx_i\beta+\lambda\alpha s_i+\epsilon_i$. By allowing $\lambda$ to vary, we can control how important the low-dimensional manifold is in predicting the outcome, as shown on the left of Figure \ref{fig:distribution}. SAEs are largely ineffective when the low dimensional manifold is largely irrelevant for predictions (small values of $\lambda$). However, the performance dramatically improves as the manifold becomes increasingly influential (large values of $\lambda$). Models that place high value on the reconstruction loss are more effective at taking advantage of this low-dimensional structure (high values of $\mu$).

To demonstrate the second claim we generated data $\bfx\in\R^{300}$ with zero mean and unit variance such that $\bfx=(1-\lambda)s_{i1}\bfw_1+\lambda s_{i2}\bfw_2+\mathbf{\eta}_i$. The outcome $y\in\R$ was generated as $y_i=s_{i2}+\epsilon_i$, using exclusively the second latent factor. By varying $\lambda$, we can explore the importance of having predictive information align with high variance components in the predictors, as is shown on the right of Figure \ref{fig:distribution}. The models that place high value on the reconstruction (larger values of $\mu$) largely ignore the predictive information when it is correlated with little variance in the predictors, while the model that emphasizes prediction overfits (smaller values of $\mu$). However, with slight increases in predictive variance, the model with small values of $\mu$ is quicker to incorporate information relevant to the outcome. Finally, when most variance in the predictors aligns with the predictive component, the models that place higher emphasis on reconstruction exploit the regularization most effectively and have the lowest generalization error.

The sparsity-in-dimensionality assumption of SAEs aligns better with network-based neuroscience than an alternative sparsity assumption in terms of the predictors \citep{Tibshirani1996}. Many of our predictors are highly correlated, such as power in adjacent frequency bins in the same brain region. It is likely that both predictors will be similarly predictive of the outcome. LASSO tends to shrink the coefficient for the slightly less useful predictor to 0. An SAE would ideally find the network responsible for the variance observed in both predictors and relate this latent network to the outcome. Given the strong evidence in favor of network-based neuroscience and previous experiments demonstrating the relevance of our analyzed features, it is reasonable to expect that SAEs will yield improved predictive performance relative to purely predictive or joint models when the relevant networks are moderately-to-highly expressed, but are potentially detrimental when the network has comparatively low expression. 

\subsection{How does Predictive Sparsity Impact SAEs?}

The SAE approach as previously defined only ensures a predictive latent subspace. However, our applications require that a single predictive network be learned while the remaining networks model less relevant dynamics. In related settings, it is common to place a hard sparsity constraint on the parameters $\Phi$, limiting the predictive influence to a single (or small number) of factors \citep{Gallagher2017}. This can lead to undesirable consequences when minimizing the loss in Equation \ref{eq:sae_loss} when the decoder involves a difficult-to-optimize objective \citep{ulrich2015gp}. In particular, one locally optimal solution is to use the predictive factors to highly overfit the outcome while the associated loadings vanish. This corresponds to a latent space factorized into a predictive and generative model; the predictive factor has no biological significance as it explains none of the dynamics, while the unpredictive factors model the electrophysiology with one-fewer latent dimensions. This local optimum is not apparent with the linear models due to more stable training properties. However, with more complex models, such as the Gaussian process model of \cite{Gallagher2017}, this is a common problem.

We address this issue using an SAE approach by replacing the penalization terms on the loadings and the latent factors with a normalization constraint on the loadings {\color{black} for each factor}. Let $\Theta_l$ be the loadings associated with factor $l$. Then, our objective function corresponding to Equation (\ref{eq:sae_loss}) is replaced with an objective function
\begin{equation}\label{eq:normalized}
    \begin{split}
        \min_{\Theta,\Phi,\calA} & \sum_{i=1}^N \mu\, \loss(\bfx_i,\hat{\bfx}_i) + \loss(y_i,\hat{y}_i) + \loss(\Phi)\\
        \text{s.t.} &\norm{\Theta_l}^2=c\\
        &\Theta\ge0,\calA(\bfx)\ge0,
    \end{split}
\end{equation}
{\color{black} where the non-negativity constraints are applied elementwise. For convenience we chose $c=\sqrt{p}$ for each factor. However the choice is arbitrary; latent variable models inherently lack multiplicative identifiability. If the latent variables are multiplied by a constant $k$ and the loadings by $1/k$, the new reconstruction $kA(\bfx)\Theta/k=A(\bfx)\Theta$ will remain unchanged.  This is not an issue for selecting a stimulation target, as we choose targets based on relative importance. Incorporating these constraints is straightforward with modern packages.  Specifically, we define hidden unconstrained variables $\Xi$, and then set the loadings as a transformation of these hidden variables $\Theta=h(\Xi)$ such that $h(\Xi)$ fulfills the constraint.  In our case, we used the softplus function, $\text{softplus}(x)=\log(1+\exp(x))$, which smoothly maps from the full real line to the positive reals. With automatic differentiation, the objective in~\eqref{eq:normalized} can be learned with respect to $\Phi$, $\Xi$, and $\calA$, rather than optimizing the loadings directly.}

In SAEs with deep decoders, implementing this type of identifiability constraint is not straightforward. However, our application requires that the SAE correspond to a joint model. Thus, the decoder corresponds to a shallow network, and there are natural normalization constraints on the associated loadings. In this model, a natural constraint on the NMF loadings is to require that the norm of each factor be equal to a constant.

This normalization constraint also possesses a scientific justification. As a single factor is used to make predictions, we can choose targetable features based on the importance of the supervised factor loadings relative to the other factor loadings (e.g., the features uniquely important to the supervised network). This relevance is evaluated by dividing each entry in the supervised factor loadings by the sum of that entry in all factor loadings. The features with the highest ratios in the supervised network are potential candidates for stimulation. 
Additionally, as the network associated with our target behavior is often relatively small in variance (hence the initial need for SAEs),  normalizing the loadings for each factor to a constant provides a convenient method to ensure that the supervised network is not deemed irrelevant due to smaller loadings, as would be the case with penalization-based identifiability constraints.


\section{Validation of our Approach with Synthetic Data}

We now provide two synthetic examples that demonstrate that our SAE-based approach can recover a single predictive factor that is robust to misspecification. In the first example, we generate data from a known NMF model and show that our SAE-based approach can recover a predictive factor under misspecification in the latent dimensionality. In the second demonstration, we generate synthetic LFP dynamics using an alternative latent variable model \citep{Gallagher2017} and extract a predictive factor using our NMF model. The first example validates two aspects; first, that the estimated factor is predictive and second, that the estimated loadings for the SAE match the true loadings. The second example validates that our NMF approach can robustly extract predictive factors in a realistic simulation of experimental conditions. In both examples, we show that alternative approaches (cutting the feedback and supervised dictionary learning) fail under our experimental conditions.

\subsection{Recovering a Synthetic NMF Component}

We generated synthetic data such that the latent factors and the conditional distribution of the observations given the factors had truncated normal distributions. This aligns with the likelihood assumed with an $L_2$ loss and non-negativity constraints. We chose a latent dimensionality of 10 and a 100-dimensional observation space to match the assumption that the number of networks is substantially lower than the observed dimensionality. The latent factors were independent with distinct variances. The outcome came from a Bernoulli distribution where the probability depended exclusively on the lowest-variance component. The data generation process can be written as
\begin{equation}\label{eq:synthetic}
    \begin{split}
        \mathbf{s}_i &\sim TN(\mathbf{0}_{10},\Sigma,0,\infty)\\
        \mathbf{x}_i|\mathbf{s}_i &\sim TN(W\mathbf{s}_i,\sigma^2_x I_{100},0,\infty)\\
        p_i &= logit^{-1}(s_{i1}-c)\\
        y_i|p_i &\sim Bern(p_i)
    \end{split}
\end{equation}

We fit an NMF model with 5 components to match the assumption that the number of estimated factors is smaller than the true number of networks. We evaluated three methods for inferring the model, cutting the feedback, joint modelling, and our SAE based approach. For each model we maintained identical identifiability constraints on the loadings corresponding to \eqref{eq:normalized}. We measured the predictive performance of the learned factor using the AUC of a test set and quantified the accuracy of the learned loadings as the cosine similarity between the true and estimated loadings. The cosine similarity between two vectors $\bfv_1$ and $\bfv_2$ is defined as 

\begin{equation}\label{eq:synthetic2}
\text{similarity}=\frac{\bfv_1\cdot \bfv_2}{\norm{\bfv_1}\norm{\bfv_2}}
\end{equation}

This quantity measures how well the two vectors align with 1 indicating perfect alignment and 0 indicating orthogonality. The cosine similarity allows us to answer how similar different brain networks are. Highly similar networks will involve similar regions with similar covariates. As such, they will largely align with cosine similarities close to 1. On the other hand, networks that involve different covariates will have lower cosine similarities. {\color{black} The permutative non-identifiability is accounted for in the SAE and joint models by the fact that a single component is supervised, so similarity is measured between the supervised loadings and the true loadings. In the joint NMF, this similarity is computed between the loadings associated with the most predictive factor and the true loadings.}

  
          \begin{figure}[H]
          \centering
              \includegraphics[width=.6\linewidth]{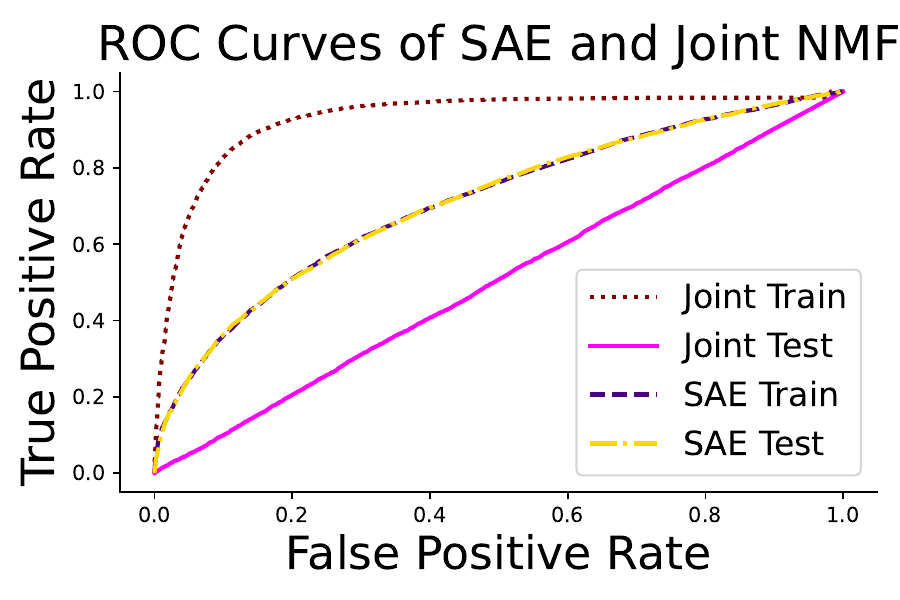}
              \caption{ \label{fig:roc} The ROC curves on the training and testing sets for the estimated joint model and SAE.}
          \end{figure}

          

\begin{table}
\caption{\label{table:syn} Prediction and Reconstruction}
\begin{tabular}{l| c  c}
Model & AUC & Similarity\\
\hline
True & $0.76$ & $1.0$ \\
Logistic & $0.71$ & $-$ \\
Cut NMF & $0.53$ & $0.76$ \\
Joint NMF & $0.50$ & $0.76$ \\
SAE NMF & $0.71$ & $0.94$ 
\end{tabular}
\end{table}



Figure \ref{fig:roc} and Table \ref{table:syn}
show the results of the predictive AUC of the single estimated predictive factor, along with the cosine similarity between the estimated loadings and the true network characteristics. For comparison we have included the values for the true model and logistic regression. The latter provides an estimate of the predictive ability of a standard linear model. Cosine similarity ranges from $0$ to $1$, and for the specific generated dataset the cosine similarity between two randomly learned loadings is $0.64$. Values larger than this indicate that the model successfully incorporated information from the true model into the estimated network.

The factor estimated via cutting the feedback contains no predictive information and does not align with the true network. This is unsurprising, as none of the estimated factors using reconstructive loss alone will contain information related to a low-variance factor as defined. More interestingly, the joint factor model provides no benefits over a cut model. This seems surprising as incorporating information related to the outcome during estimation \textit{should} improve predictive performance. However, the figure on the right shows why this is not the case. The AUC on the training set of the joint model is far higher than the SAE, and higher than the theoretical bound provided by the true model. However, predictive ability returns to random chance when the factors are estimated in the test set without knowledge of the outcome. This is the ``factor dragging'' problem stated in Section 2. The model overfits the predictive factors without modifying the corresponding loadings, making prediction dependent on knowledge of the outcome. Our SAE approach avoids this overfitting and accurately characterizes the network.

\subsection{Extracting Predictive Features in Synthetic LFPs}

We now validate our SAE-NMF approach using synthetic LFPs to match our experimental conditions as closely as possible. The data were generated using a previously-developed factor model specifically designed for analyzing LFPs, referred to as Cross-Spectral Factor Analysis \citep{Gallagher2017}. These models use Gaussian processes in a multiple kernel learning framework to represent the spectral features in a low-dimensional space of latent factors. This approach functions as a generative model given a prior on the latent factors.

\begin{figure}[H]
          \centering
              \includegraphics[width=.6\linewidth]{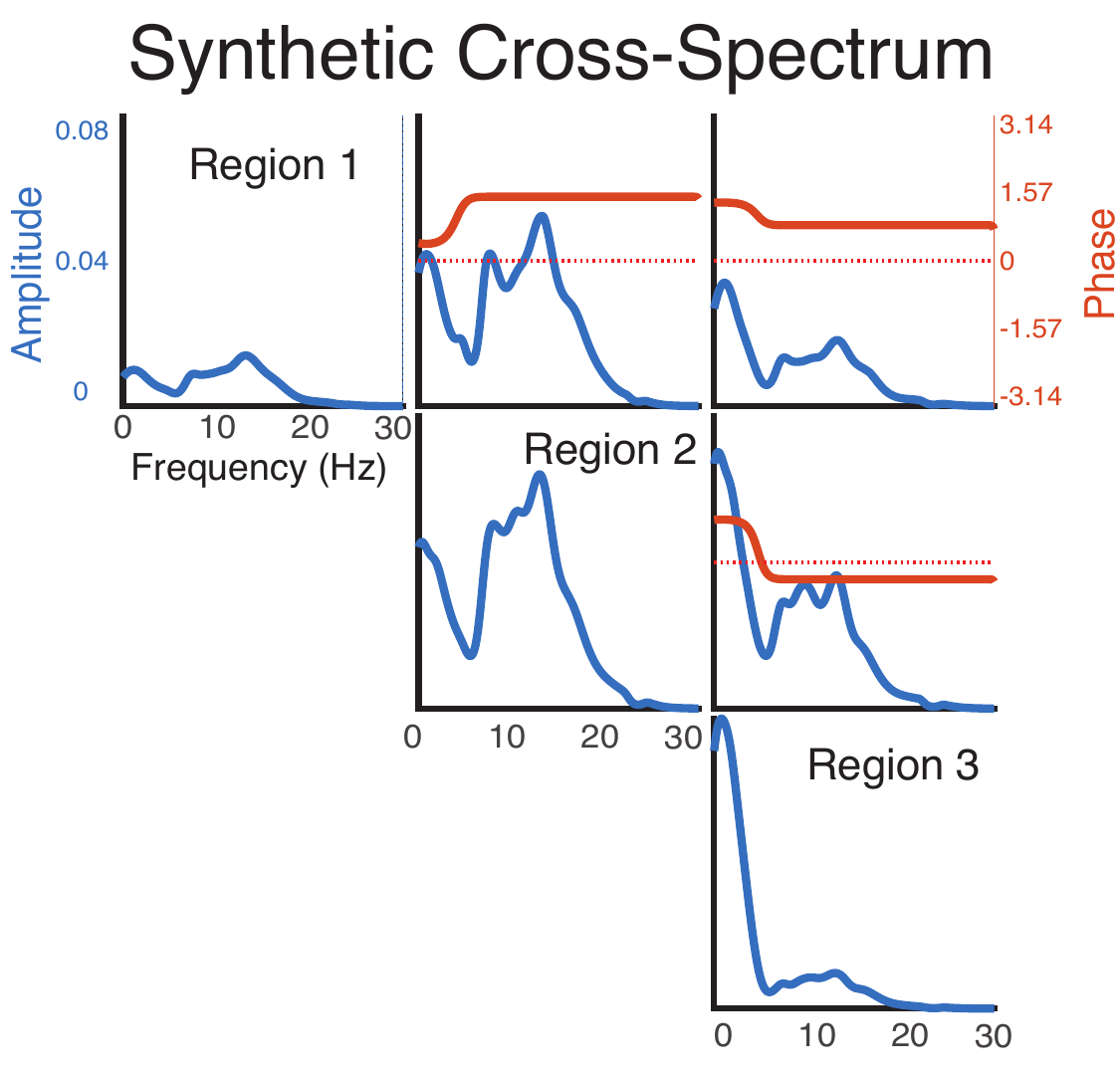}
              \caption{ \label{fig:synth_CSFA} The relative importance of the predictive CSFA factor in 3 of the 8 generated regions along with phase offset (indicates directionality).}
          \end{figure}

We initialized a CSFA model with 30 latent components and 8 measured brain regions and generated 20,000 samples of  synthetic LFPs using the sampling method described in Section A.6. The draws associated with a single latent factor were used to generate a synthetic binary outcome. The associated power spectrum of a subset of the regions is plotted in Figure ~\ref{fig:synth_CSFA}. This dataset reflects many of our assumptions of brain dynamics; the latent variables have a substantial amount of sparsity, a small number of latent variables are responsible for the outcome of interest, and the spectral features associated with the latent variable are sufficient to characterize the brain network.

We then calculated the power and coherence features of the synthetic data using the same procedure described in Section 2. This allows us to examine the performance of our approach under realistic circumstances.  We compare our supervised approach to a logistic regression model using the observed features, a sequentially fit NMF model, and a joint model. Each NMF model was estimated with only 5 latent factors instead of the 30 factors used to generate the data. Therefore, our example demonstrates the effect of misspecification, not only in the latent dimensionality, but also in the observational likelihood of the data under realistic assumptions.

   \begin{table}
   \caption{\label{table:syn2} Prediction Using Synthetic LFPs}
   \begin{tabular}{l| c }
Model & AUC \\
\hline
Logistic & $0.58$ \\
Cut NMF & $0.50$\\
Joint NMF & $0.50$ \\
SAE NMF & $0.59$
\end{tabular}


\end{table} 

The results are shown in Table~\ref{table:syn2}. The cut model fails to capture any relevant information to the outcome, which we attribute to the substantial misspecification. Correspondingly, the joint model also fails to yield accurate predictions as it predicts nearly perfectly on the training data but does not incorporate that information into the associated loadings. However, our SAE approach is able to discover relevant features for prediction. In fact, it slightly improves on the pure logistic regression model aligning with the theoretical results of \citet{Le2018}. 

The results of these two synthetic examples suggest that our approach will be able to accurately recover the relevant dynamics associated with the outcome of interest and be robust to the types of misspecification that can be reasonably expected when analyzing LFPs.

\section{Estimating Networks Associated with Stress and Bipolar Disorder}

We have demonstrated why and how our SAE approach can yield improved predictive performance in synthetic LFP data.  Furthermore, our approach can accurately estimate the associated network characteristics in a known model and handle misspecification of both the latent distribution and observational likelihoods.  We now demonstrate the predictive abilities of SAEs on the TST dataset. We use an NMF model with a single supervised factor to force all the predictive information into a single brain network. Our objective is to find a single network predictive of stress and a single network predictive of genotype. We fit a model with 10 factors and a set supervision strength of $\mu=1$. The complete details of the implementation and the features are provided in the Appendix. 

We compared the results from our approach to the two most relevant competitors, a joint factor model with a single supervised factor and ``cutting the feedback'' with different numbers of latent factors. Since our objective is to find a \textit{single} predictive network, we tested two measures of predictive ability when comparing to cutting the feedback. The first was the predictive ability using the entire latent subspace and the second was the predictive ability using the single most predictive factor on the training set. One method for improving the predictive performance with cutting the feedback approaches is to increase the number of latent factors, so we compare the performance of a $25$-factor model in addition to the corresponding $10$-factor model. We also compared the reconstructive losses to ensure biological relevance of the learned model.

Table \ref{tab:task} shows the predictive and reconstructive metrics using all stated models for predicting stress, along with  $95$ \% confidence intervals. It is apparent that our SAE-based approach is substantially better than cutting the feedback or fitting a joint factor model in predictive ability. Our approach with a single factor achieves an AUC of $0.94$ relative to $0.76$ for both the cut model and joint model. Even with more factors, a cut model approach still yields worse predictions. In fact, increasing the latent dimensionality can make the prediction using a single factor worse as the relevant information becomes divided between an increasing number of factors. The reconstruction losses were better for the cut and joint models at $0.043$ and $0.045$ respectively, which is unsurprising as the encoder constrains how well the SAE can adapt the factors to each observation. However, the SAE still explains a substantial portion of the variance with a reconstruction loss of $0.051$ compared to $0.09$ loss from using the mean, indicating that the estimated factors are still biologically relevant.

\begin{table}
\caption{\label{tab:task} Prediction of Stress}
\begin{tabular}{l| c  c  c}
Model & Prediction: 1 Factor & Prediction: All Factors & Reconstruction \\\hline
NMF--10 Components & $0.76\pm0.03$ & $0.83\pm0.02$ & $0.043$ \\
NMF--25 Components & $0.69\pm0.04$ & $0.87\pm0.05$ & $0.033$ \\
NMF--Joint Model & $0.76\pm0.01$ & $0.76\pm0.01$ & $0.045$ \\
\textbf{NMF--SAE} & $\mathbf{0.94\pm0.01}$ & $\mathbf{0.94\pm0.01}$ & $0.051$ 
\end{tabular}
\end{table}

{\color{black} 

Choosing the optimal number of factors is a challenging problem. We note that several approaches exist to select the dimensionality in factor models, but they do not naturally apply in our context. For example, there are a number of Bayesian hierarchical methods that define priors for helping choose the dimensionality \citep{Bhattacharya2011} or penalization-based schemes to balance generative capability with complexity \citep{minka2000automatic}. It is relatively common in machine learning to choose the number of components based on heuristics of explained variance \citep{ferre1995selection} or by choosing the number of components through a cross-validation procedure. However, it is non-trivial to extend these existing techniques to the needs of our application, as it is challenging to include the hierarchical priors in the SVAE formulation, and a focus exclusively on prediction does not lend itself to finding a system that reconstructs the data well and can be explained as networks. Instead, for this application we chose 10 factors to match previously learned models \citep{Gallagher2017}. This allowed for the generative factors to account for the baseline variance while not rendering the predictive factor irrelevant in the reconstruction of the electrophysiology.}

\begin{figure}[t!]
\centering
\includegraphics[width=0.6\textwidth]{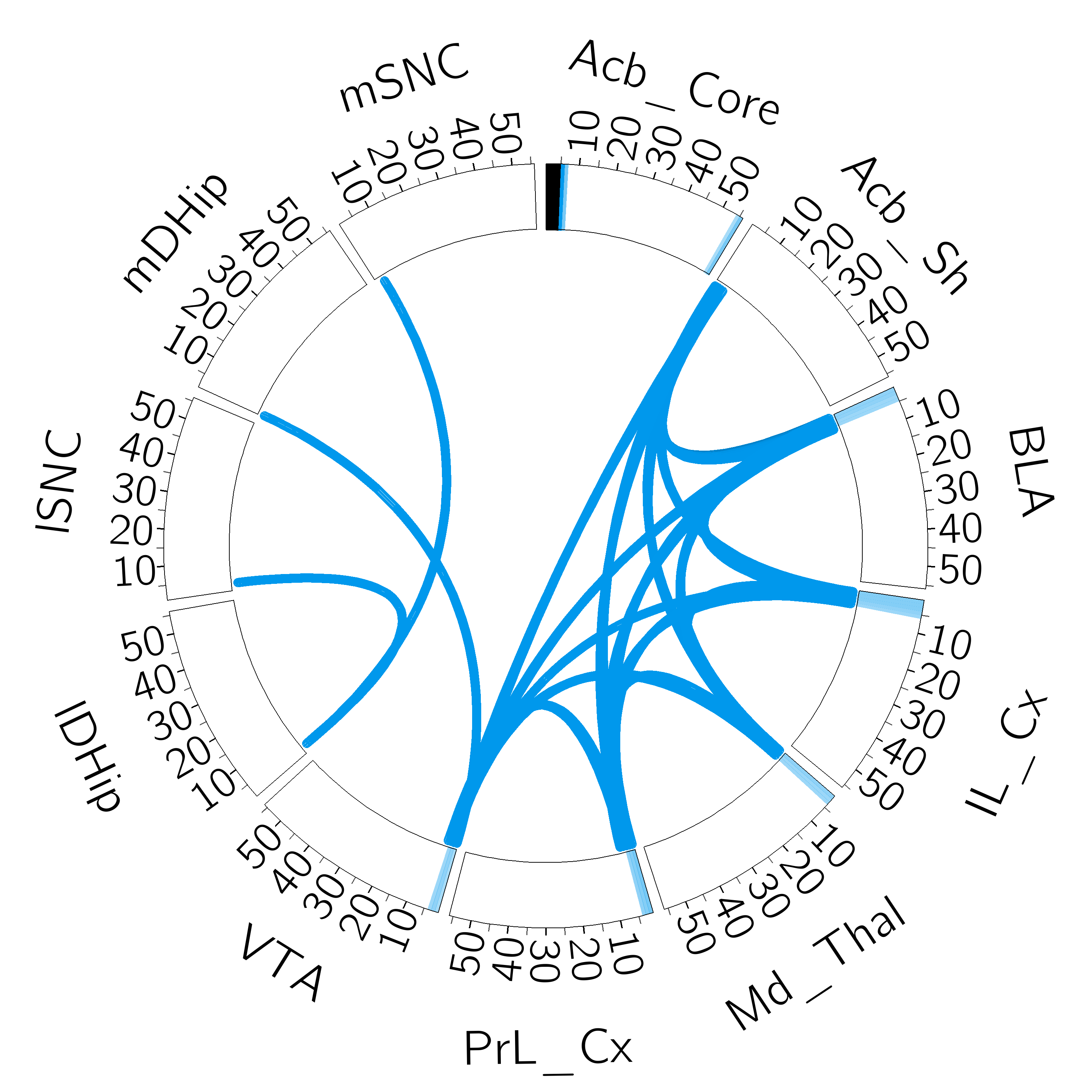}
\caption{\label{fig:task} Visualization of the brain network found to be predictive of experimental condition. We first normalize the loadings of the specific factor relative to the loadings of all factors. These values evaluate how much influence the network had on a specific feature. We then thresholded for the largest coefficients; any covariate that exceeded this threshold was included. Each segment of the circle represents a different brain region, and the numbers $(10,\dots,50)$ represent frequencies in Hertz. Any power features that exceeded the threshold were included as a band on the rim. Any coherence features were plotted as a ``spoke''. As an example interpretation, this network substantially influenced the power in the prefrontal cortex (PrL\_Cx and IL\_Cx) at low frequencies, as well as the coherence between the prefrontal cortex and the acumbens, amygdala, thalmus, and ventral tegmental area (Acb\_Sh, BLA, Md\_Thal, and VTA).}
\end{figure}

The sign of the predictive model associated with the supervised factor in the SAE was consistently negative across all the data splits. This indicates that the learned network was less active in stressful conditions as compared to a non-stressful environment. Figure ~\ref{fig:task} shows one method for visualizing the network. Each segment along the edge represents power in the labeled brain region. The ``spokes'' in the circle represent coherence between the two specified regions. {\color{black} Stimulation targets are chosen as region/frequency combinations that are particularly influential in the supervised network. These influential combinations will have large loadings in spectral power and coherences between that region and other brain regions. By stimulating this ``central'' region, we can influence the entire network and its associated behavior \citep{mague2022brain}.}

In this particular case, the network negatively associated with stress is characterized by power in the low frequencies of $1-4$ Hz in the prefrontal cortex (IL\_Cx and PrL\_Cx), thalamus (Md\_thal), amygdala (BLA), and ventral tegmental area (VTA) along with high levels of coherence between these regions. It also is characterized by coherence at these frequencies within the hippocampus (mDHipp and IDHipp). Since this network is lower in the stressful situation, stimulation in the prefrontal cortex or VTA at low frequencies should modify the dynamics in a stressful situation to more closely align with the dynamics observed in non-stressful situations. Such stimulation would answer whether the network is causal, and potentially ameliorate the impact of stressful behavior. We note that a prior stimulation procedure was used that stimulated the thalamus (Md\_thal) in a phase-locked manner to $3-7$ Hz waves in the prefrontal cortex \citep{Carlson2017}.  This stimulation procedure aligns with the network discovered by our approach, and reduced time immobile in the tail suspension test, which is consistent with a reduction in stress.

Differentiating Clock-$\Delta$19 vs wild type dynamics is a far more difficult task as compared to characterizing stress. This would unsurprisingly indicate that tail suspension and non-stressful situations are more distinctive than the baseline dynamics due to modification of a single gene. Yet also here our SAE approach outperforms prediction using cut or joint models as shown in Table \ref{tab:genotype}. In this case, adding extra factors to the cut model yields a closer (but still inferior) predictive ability to the SAE model. However, the previous task showed that a strategy relying on increasing latent dimensionality to improve predictions with a single factor will give inconsistent results. Our SAE approach provides a more reliable method for finding a single network correlated with the outcomes of interest in neuroscience.

\begin{figure}[t]
\begin{center}
\includegraphics[width=0.6\textwidth]{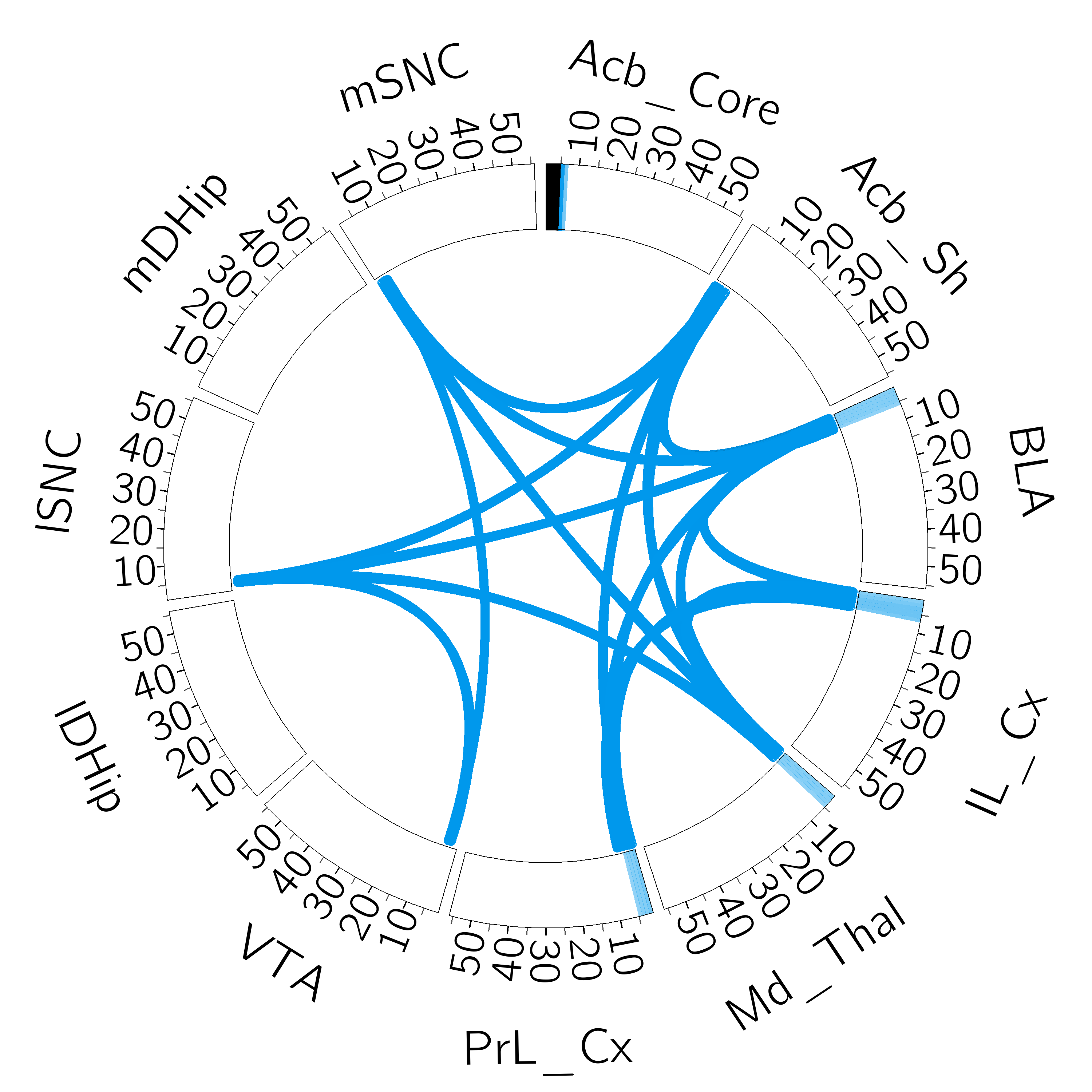}
\end{center}
\caption{\label{fig:genotype} The predictive network of the Clock-$\Delta$19 genotype, with an identical interpretation to Figure \ref{fig:task}.} 
\end{figure}

\begin{table}
\caption{\label{tab:genotype} Prediction of Genotype}
\begin{tabular}{l| c  c  c}
Model & Prediction: 1 Factor & Prediction: All Factors & Reconstruction \\\hline
NMF--10 Components & $0.56\pm0.03$ & $0.59\pm0.02$ & $0.043$ \\
NMF--25 Components & $0.64\pm0.04$ & $0.67\pm0.05$ & $0.033$ \\
NMF--Joint Model & $0.55\pm0.01$ & $0.55\pm0.01$ & $0.046$ \\
\textbf{NMF--SAE} & $\mathbf{0.68\pm0.04}$ & $\mathbf{0.68\pm0.04}$ & $0.055$ 
\end{tabular}
\end{table}
The network found in each of the splits was consistently negatively correlated with the Clock-$\Delta$19 genotype and the network found in one of the splits is shown in Figure \ref{fig:genotype}. That this network is negatively associated with Clock-$\Delta$19 indicates that this stimulation in Clock-$\Delta$19 mice would make their dynamics more closely align with the wild type population. This network is characterized by power in the prefrontal cortex at low frequencies similar to the stress network. However, this network is weaker in the hippocampus (IDHip and mDHip). This would indicate that a reasonable location for stimulation is the prefrontal cortex at low frequencies. 

Our emphasis in this work has been finding networks that can be used for stimulation, and therefore we have largely refrained from interpreting the scientific conclusions derived from the estimated networks in the TST dataset. Nevertheless, it is easy to see the similarities between the estimated stress network and the estimated Clock-$\Delta$19 network. In fact, the cosine similarity between the networks in Figures 5 and 6 is 0.99, whereas if we pick a random network in the task model and a random network in the genotype model, the average cosine similarity is 0.38. Based on this, our models would suggest that the differences in genotype are similar to the observed differences induced by stressful and non-stressful conditions. It is encouraging to note that this observation has matched previous scientific experimentation (Murata et al., 2005).  Additionally, this implies that the stress felt by the Clock-$\Delta$19 animals is lower than WT animals, which aligns with the finding that they are more active during the TST (Carlson et al., 2017). We note that these findings show an appealing aspect of our approach: by using an interpretable and biologically relevant generative model in our SAE, we have been able to scientifically analyze the differences between the experimental groups.

\begin{figure}[t!]
\begin{center}
\includegraphics[width=0.6\textwidth]{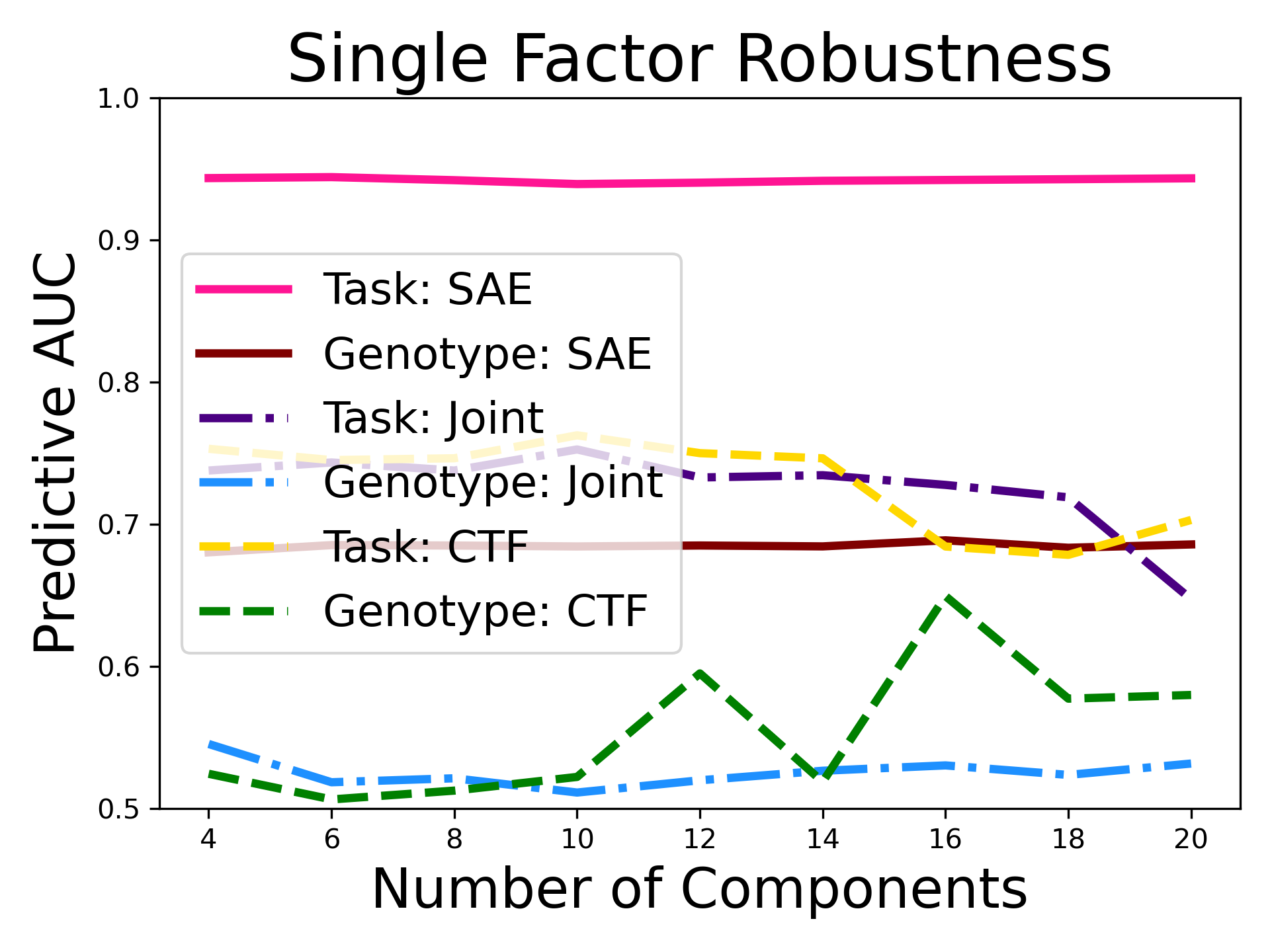}
\end{center}
\caption{{\color{black}\label{fig:dimennsionality} The impact of latent dimensionality on the SAE predictions of task. SAE refers to our supervised autoencoder approach while CTF refers to the ``cutting-the-feedback'' approach.}}
\end{figure}

{\color{black} As a final comparison, we can also empirically verify our claim that SAEs predictive ability is robust to the latent dimensionality. Figure \ref{fig:dimennsionality} shows the predictive ability of our SAE-based approach, a SDL approach, and a ``cutting-the-feedback'' as a function of latent dimensionality, predicting both genotype and stress condition using the most predictive factor. These results were obtained evaluating the general population rather than a mouse-by-mouse basis to match standard statistical evaluation techniques. We can see on both tasks that the SAE maintains a consistently-high predictive ability with all latent dimensionalities. However, the other two methods not only fail to match this predictive ability, but also fluctuate rather than providing a clear monotonic trend. This illustrates a substantial problem that has traditionally plagued network-based neuroscience; even simple parameter choices can cause dramatic shifts in predictive results.

}

\section{Conclusions}

We have developed an approach for finding a single stimulatable brain network that correlates with a condition or behavior. Our strategy is based on supervised autoencoders, which are more conducive to finding a single network predictive of an outcome of interest than joint factor models. We are able to find a brain network that is correlated with stress and a network that is correlated with a genotype linked to bipolar disorder. This approach naturally leads to candidate brain regions for stimulation to modify the network.

While the objective of this work is developing an approach for identifying relevant brain networks, we also provide broad conceptual contributions to understand why, how, and when supervised autoencoders are effective model choices. This is done by deriving supervised autoencoders in a manner that elucidates the source of superior predictive ability relative to joint factor models, and using analytic solutions in a simple case to illustrate why carefully designed SAEs are useful in our applications.

\section{Acknowledgements}

We would like to thank Michael Hunter Klein and Neil Gallagher for providing helpful feedback and testing the code for the NMF and PCA models. Research reported in this publication was supported by the National Institute of Biomedical Imaging and Bioengineering and the National Institute of Mental Health through the National Institutes of Health BRAIN Initiative under Award Number R01EB026937 to DC and KD and by a WM Keck Foundation grant to KD. DD was funded by National Institute of Health grants R01ES027498 and R01MH118927.

\bibliographystyle{rss}
\bibliography{jrsc}

\clearpage

\appendix
\section{Appendix}

\subsection{Derivation of an SAE via Lagrangian Relaxation}

We denote $\{\bfx_i\}_{i=1,\dots,N}\in\calX$ as the measured dynamics (predictors) and $\{y_i\}_{i=1,\dots,N}\in\calY$ as the corresponding trait or outcome. The latent factors are denoted $\{\bfs_i\}_{i=1,\dots,N}\in\mathbb{R}^L$, $\Theta$ are the parameters relating the factors to $\bfx$, and $\Phi$ are the parameters relating the factors to $y$. Let us define the predictive feedforward network as $\hat{y}=f_{\Phi}(\calA(\bfx))$, where $\calA:\bfX\rightarrow\mathbb{R}_+^{L}$ is the encoder, and the reconstructive autoencoder is a composition of the encoder and a decoder $\hat{\bfx}=g_\Theta(\calA(\bfx))$. Then the empirical risk minimization of the predictive loss yields an objective function

\begin{equation}
    \min_{\Phi,\calA}\sum_{i=1}^N \loss(y_i,\hat{y}_i) + \loss(\calA(\bfx_i)) + \loss(\Phi)
\end{equation}

Adding a constraint on the reconstructive loss in a manner corresponding to structural risk minimization yields
\begin{equation}
    \begin{split}
        \min_{\Phi,\calA} &\sum_{i=1}^N \loss(y_i,\hat{y}_i) + \loss(\calA(\bfx_i)) + \loss(\Phi)\\
        \text{s.t.} &\min_{\Theta}\sum_{i=1}^N \loss(\bfx_i,\hat{\bfx}_i) + \loss(\Theta)<L
    \end{split}
\end{equation}

This constraint on the reconstruction loss is difficult in practice. To circumvent this difficulty, we introduce a Lagrange multiplier $\mu$ and the unconstrained version becomes
\begin{equation}
    \min_{\Phi,\calA}\sum_{i=1}^N \loss(y_i,\hat{y}_i) + \loss(\calA(\bfx_i)) + \loss(\Phi) - \mu(L-\sum_{i=1}^N \loss(\bfx_i,\hat{\bfx}_i) + \loss(\Theta))
\end{equation}
Since L does not depend on the parameters we then treat $\mu$ as a tuning parameter controlling the strength of the supervision loss. This yields the form of Equation (\ref{eq:sae_loss}).

\subsection{Derivation of SAE and Joint Factor Model Analytic Solutions with an $L_2$ Loss}

We assume that both the predictors and the outcome are demeaned. For convenience, we will define matrix forms of the data and factors: $X=[\bfx_1,\dots,\bfx_N]\in\mathbb{R}^{p\times N}$, $Y=[\bfy_1,\dots,\bfy_N]\in\mathbb{R}^{q\times N}$, and $S=[\bfs_1,\dots,\bfs_N]\in\mathbb{R}^{L\times N}$. We let the model parameters $\Theta=W$ and $\Phi=D$ for matrices $W\in\mathbb{R}^{p\times L}$ and $D\in\mathbb{q\times L}$. For the associated SAE model, we assume a linear encoder, $\calA(\bfx)=A\bfx$. Both replace the negative log likelihood with an $L_2$ loss. Rather than penalizing the factors and loadings we placed a normalization constraint on the loadings, analogous to PCA. 

\subsection{Deriving the Joint Model Form}

With simplifications, the objective corresponding to Equation (\ref{eq:jfm}) is
\begin{equation}
    \min_{W,D,\{\bfs_i\}_{i=1,\dots,N},W^TW=I}\sum_{i=1}^N\norm{\bfx_i-W\bfs_i}_F^2 + \mu\norm{\bfy_i-D\bfs_i}_F^2.
\end{equation}
We can rewrite the above objective matrix form as
\begin{equation}
    \min_{W,D,S,W^TW=I}\norm{X-WS}_F^2 + \mu\norm{Y-DS}_F^2.
\end{equation}
This is equivalent to the trace representation of 
\begin{equation}
\begin{split}
    \min_{W,D,S,W^TW=I}&\text{Tr}((X-WS)^T(X-WS)) + \mu\text{Tr}((Y-DS)^T(Y-DS))\\
    =\min_{W,D,S,W^TW=I}&\text{Tr}(X^TX-2X^TWS+S^TW^TWS) + \mu\text{Tr}(Y^TY-2Y^TDS+S^TD^TDS)
    \end{split}
\end{equation}

We can now take the partial derivatives with respect to $W$, $S$, and $D$ using standard techniques with the aim to find the fixed points where each partial derivative vanishes. Denoting the previous loss as $F$, the first partial derivative with respect to $W$ is 
\begin{equation}
    \begin{split}
        0&=\frac{\partial F}{\partial W}=-2XS^T+2WSS^T\\
        2XS^T &= 2WSS^T\\
        W &= XS^T(SS^T)^{-1}.
    \end{split}
\end{equation}
The derivative with respect to $D$ is
\begin{equation}
    \begin{split}
        0&=\frac{\partial F}{\partial D}=-2YS^T+2DSS^T\\
        2YS^T &= 2DSS^T\\
        D &= YS^T(SS^T)^{-1}.
    \end{split}
\end{equation}
And the derivative with respect to $S$ is
\begin{equation}
    \begin{split}
        0&=\frac{\partial F}{\partial S}=-2W^TX+2W^TWS-2\mu D^TY + 2\mu D^TDS\\
        W^TWS+\mu D^TDS &= \mu D^TY + 2 W^TX\\
        S &= (W^TW+\mu D^TD)^{-1}(W^TX+\mu D^TY).
    \end{split}
\end{equation}
These conditions will all be satisfied at the fixed point solution to this problem. We will first solve for a single latent feature, so the matrices $W$, $D$, and $S$ are vectors $\bfw$, $\bfd$, and $\bfs$. With that we can rewrite the above condition as 
\begin{equation}
\bfs=(\norm{\bfw}^2+\mu\norm{\bfd}^2)^{-1}(\bfw^TX+\mu \bfd^T Y).
\end{equation}
We deonte $\alpha=(\norm{\bfw}^2+\mu\norm{\bfd}^2)^{-1}$ and $\gamma=(\bfs\bfs^T)^{-1}$. Since $\bfs$ is a vector, both $\alpha$ and $\gamma$ are scalars in this case. Thus, in a single feature case, the fixed point equations can be written as
\begin{equation}
    \begin{split}
        \bfw &= \gamma X\bfs^T\\
        \bfd &= \gamma Y\bfs^T\\
        \bfs &= \alpha (\bfw^TX+\mu \bfd^TY)
    \end{split}
\end{equation}
When we substitute $\bfs$ into the previous equations we end up with
\begin{equation}
    \begin{split}
        \bfw &= \alpha\gamma X(X^T\bfw+\mu Y^T\bfd)\\
        \bfd &= \alpha\gamma Y(X^T\bfw+\mu Y^T\bfd)
    \end{split}
\end{equation}
Because $\alpha$ and $\gamma$ are scalars we note that the solution to the equations above must be an eigenvector of 
\begin{equation}
    B = \begin{bmatrix}
\mu XX^T & XY^T \\
\mu YX^T & YY^T
\end{bmatrix},
\end{equation}
where the first part of the eigenvector corresponds to the solution of $\bfw$ and the second part corresponds to the solution of $\bfd$. In practice, we have always found that the solution corresponds to the largest eigenvector. Finding further dimensions can be done iteratively by subtracting out the previous variance explained then repeating the procedure. In practice, we have found these to be equivalent to the $L$ largest eigenvalues of $B$, which is supported by the fact that when $\mu=1$ this yields the classic PCA solution.

\subsection{Deriving the SAE Form}

We similarly rewrite the objective of an SAE in matrix form as
\begin{equation}
    \min_{W,D,A,W^TW=I}\norm{X-WAX}_F^2 + \mu\norm{Y-DAX}_F^2.
\end{equation}
This is equivalent to the trace representation of 
\begin{equation}
\begin{split}
    \min_{W,D,A,W^TW=I}&\text{Tr}((X-WAX)^T(X-WAX))+\mu \text{Tr}((Y-DAX)^T(Y-DAX))\\
    =\min_{W,D,A,W^TW=I}&\text{Tr}((X-WAX)^T(X-WAX))+\mu \text{Tr}((Y-DAX)^T(Y-DAX))
    \end{split}
\end{equation}

We can now take the partial derivatives with respect to $W$, $A$, and $D$ using standard techniques with the aim to find the fixed points where each partial derivative vanishes. Denoting the previous loss as $F$, the first partial derivative with respect to $W$ is 
\begin{equation}
    \begin{split}
        0&=\frac{\partial F}{\partial W}=-2XX^TA^T+2WAXX^TA^T\\
        2XX^TA^T &= 2WAXX^TA^T\\
        W &= XX^TA^T(AXX^TA^T)^{-1}.
    \end{split}
\end{equation}
The derivative with respect to $D$ is
\begin{equation}
    \begin{split}
        0&=\frac{\partial F}{\partial D}=-2YX^TA^T+2DAXX^TA^T\\
        2YX^TA^T&=2DAXX^TA^T\\
        D &= YX^TA^T(AXX^TA^T)^{-1}.
    \end{split}
\end{equation}
And the derivative with respect to $S$ is
\begin{equation}
    \begin{split}
        0&=\frac{\partial F}{\partial A}=-2W^TXX^T+2W^TWAXX^T-2\mu D^TYX^T+2\mu D^TDAXX^T\\
        W^TWAXX^T+\mu D^TDAXX^T &= \mu D^TYX^T + 2 W^TXX^T\\
        A &= (W^TW+\mu D^TD)^{-1}X^T(XX^T)^{-1}.
    \end{split}
\end{equation}
Once again, we will first solve for a single latent feature, so the matrices $W$, $D$, and $A$ are vectors $\bfw$, $\bfd$, and $\bfa$. With that we can rewrite the above condition as 
\begin{equation}
\bfa=(\norm{\bfw}^2+\mu\norm{\bfd}^2)^{-1}(\bfw^TX+\mu \bfd^T Y)X^T(XX^T)^{-1}.
\end{equation}

We deonte $\alpha=(\norm{\bfw}^2+\mu\norm{\bfd}^2)^{-1}$ and $\gamma=(\bfa XX^T\bfa^T)^{-1}$. Since $\bfa$ is a vector, both $\alpha$ and $\gamma$ are scalars in this case. Thus, in a single feature case, the fixed point equations can be written as
\begin{equation}
    \begin{split}
        \bfw &= \gamma XX^T\bfa^T\\
        \bfd &= \gamma YX^T\bfa^T\\
        \bfa &= \alpha (\bfw^TX+\mu \bfd^TY)X^T(XX^T)^{-1}
    \end{split}
\end{equation}
When we substitute $\bfa$ into the previous equations we end up with
\begin{equation}
    \begin{split}
        \bfw &= \alpha\gamma XX^T(XX^T)^{-1}X(X^T\bfw+\mu Y^T\bfd)\\
        \bfd &= \alpha\gamma YX^T(XX^T)^{-1}X(X^T\bfw+\mu Y^T\bfd)
    \end{split}
\end{equation}
Because $\alpha$ and $\gamma$ are scalars we note that the solution to the equations above must be an eigenvector of 
\begin{equation}
    B = \begin{bmatrix}
\mu XX^T & XY^T \\
\mu YX^T & YP_XY^T
\end{bmatrix},
\end{equation}
where the first part of the eigenvector corresponds to the solution of $\bfw$ and the second part corresponds to the solution of $\bfd$. In practice, we have always found that the solution corresponds to the largest eigenvector. Finding further dimensions can be done iteratively by subtracting out the previous variance explained then repeating the procedure.

\subsection{Implementation of NMF Models}

The NMF models used in this paper were implemented in Tensorflow 2.0, with the code available in a public github repository. The loadings and factors were initialized using a standard NMF implemented in sklearn. The loadings were rotated so that the most predictive factor was the designated supervised factor. With the SAE version, the encoder was initialized as the coefficients of a predictive linear model that predicted the initialized factors using the covariates. This was implemented using elastic net from sklearn. The joint factor model was implemented as batch training while the SAE was learned with stochastic gradient descent. The learning rates were chosen as values that had worked previously.

\subsection{Synthetic LFP Generation}
We generated a CSFA-CSM model according to \cite{Gallagher2017} with the code located in a Github repository located at 

\url{https://github.com/neil-gallagher/CSFA}.

We initialized a CSFA-CSM model with 30 latent variables, 8 channels, and 3 spectral Gaussian mixtures using the default method. Using this model 20000 LFP windows were generated using the default sampling method, with each window containing 1000 samples over a 1 second interval. The synthetic outcome was generated using the sampled latent factors from a single latent component. These synthetic LFPs were then processed using our standard pipeline used in Section 5. The CSFA-CSM model, synthetic LFPs and outcome are included with the paper.

\clearpage

\end{document}